\documentclass[10pt,letterpaper]{article}
\usepackage[top=2.75cm,bottom=2.5cm,left=2.5cm,right=2.5cm]{geometry}
\usepackage{lmodern}
\usepackage{amsmath,amssymb,array,textcomp,graphicx,booktabs,multirow,needspace,float}
\usepackage{parskip,microtype}
\usepackage{xcolor,titlesec,caption,fancyhdr}
\definecolor{onethreeblue}{HTML}{2E5AA8}
\titleformat{\section}{\sffamily\bfseries\color{onethreeblue}\large}{\thesection}{1em}{}
\titleformat{\subsection}{\sffamily\bfseries\color{onethreeblue}}{\thesubsection}{1em}{}
\titleformat{\subsubsection}{\sffamily\bfseries\color{onethreeblue}\small}{\thesubsubsection}{1em}{}
\usepackage[numbers,sort&compress]{natbib}
\usepackage[colorlinks=true,allcolors=onethreeblue]{hyperref}
\fancypagestyle{firstpage}{\fancyhf{}\fancyhead[L]{\includegraphics[width=50mm]{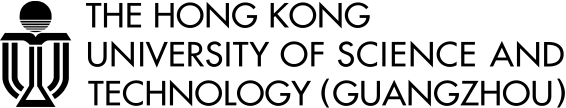}}\fancyfoot[C]{\thepage}}
\providecommand{\eg}{e.g.,}
\providecommand{\ie}{i.e.,}
\title{\sffamily\bfseries\fontsize{17}{20}\selectfont Bri3D: Bridging the Object--Scene Gap in Single-Image 3D Generation via Scene-Anchor}
\author{\small Hongye Hou\textsuperscript{1,*}, Shuai Zhang\textsuperscript{1,*}, Hongjie Li\textsuperscript{2,*}, Zhuoxiao Li\textsuperscript{1},\\
\small Haiping Wang\textsuperscript{3}, Yuan Liu\textsuperscript{3}, Wufan Zhao\textsuperscript{1,\textdagger}}
\date{}
\begin{document}
\maketitle
\thispagestyle{firstpage}
\vspace{-1.5em}
\begin{center}
\setlength{\parskip}{0pt}
\footnotesize
\textsuperscript{1}The Hong Kong University of Science and Technology (Guangzhou)\par
\textsuperscript{2}Taobao \& Tmall Group of Alibaba\par
\textsuperscript{3}The Hong Kong University of Science and Technology\par
\textsuperscript{*}Equal contribution: Hongye Hou, Shuai Zhang, and Hongjie Li.\par
\textsuperscript{\textdagger}Corresponding author: Wufan Zhao (\href{mailto:wufanzhao@hkust-gz.edu.cn}{wufanzhao@hkust-gz.edu.cn}).\par
\end{center}
\begin{figure}[H]
\centering
    \includegraphics[width=0.95\textwidth]{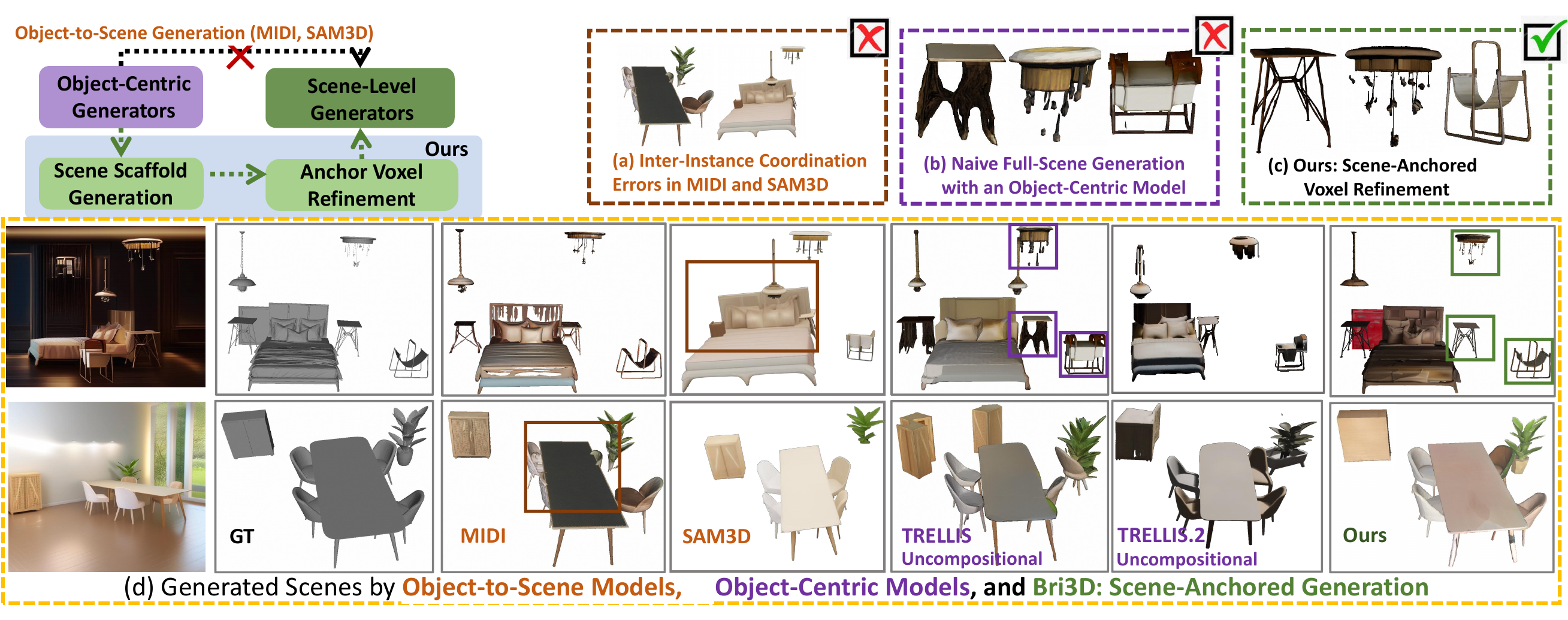}
    \caption{
    \textbf{Bridging the object--scene generation gap.}
    Object-to-scene methods, such as MIDI and SAM3D, extend object
    generation to scene composition, where coordination errors can still
    produce collisions and implausible arrangements.
    Object-centric generators produce high-fidelity individual shapes,
    but their naive extension to full scenes causes structural collapse
    and distorted geometry. In contrast, Bri3D first generates a coherent instance-aware scene scaffold and then performs scene-anchored prior refinement, producing pose-preserving voxel supports for
detail synthesis.
    }
    \label{fig:intro}

\end{figure}
\begin{abstract}
Generating compositional 3D scenes from a single image requires recovering detailed object geometry and coherent spatial arrangements. Existing object-to-scene methods, including SAM3D and MIDI, extend object generation through image-context reasoning or inter-instance modeling. Coordinating objects remains challenging with limited scene-level supervision. Directly extending object-centric generators to entire scenes can degrade geometry and fail to preserve instance scale, orientation, and pose. We propose Bri3D, a framework bridging scene-level structural reasoning and object-level generative priors through instance-aware voxel scene anchors. Bri3D predicts a color-coded scene scaffold encoding coarse object geometry and spatial arrangements. Each instance is extracted as a scene anchor and refined into a pose-preserving, prior-compatible voxel support, enabling a frozen object generator to synthesize detailed textured geometry in the anchor-defined spatial reference. This coarse-to-fine design separates scene organization from object-scale detail synthesis, focusing scene-level training on spatial organization and anchor refinement while reusing pretrained object priors. Experiments demonstrate improvements in scene layout, object geometry, visual quality, and generalization. Bri3D predicts scene layouts efficiently; full scene generation is faster than SceneGen and SAM3D.
\end{abstract}
\clearpage
\section{Introduction}
\label{sec:intro}

Generating high-fidelity 3D indoor scenes from a single image
remains a long-standing challenge at the intersection of computer
vision, graphics, and embodied AI
~\cite{nie2020total3dunderstanding,liu2022towards,chu2023buol,
dong2025hiscene,yao2025cast}.
Recent 3D generative models have achieved remarkable progress in
synthesizing realistic individual objects
~\cite{zhang2024clay,wu2024direct3d,ren2024xcube,
tochilkin2024triposr,xiang2025structured,ye2025hi3dgen,
li2025triposg,wu2025direct3d,chen2025ultra3d}.
Extending this progress to indoor scenes, however, remains difficult.
Unlike isolated object generation, scene reconstruction must jointly
recover detailed object geometry and a globally valid spatial
configuration governed by relative scale, orientation, support
relationships, and collision avoidance.

Recent object-to-scene methods combine object priors with inter-instance
modeling or image-conditioned shape and pose inference
~\cite{huang2025midi,meng2025scenegen,sam3dteam2025sam3d3dfyimages}.
They improve scene composition, yet coordinating object geometry and
placement remains challenging: errors can produce intersections,
floating furniture, and inconsistent arrangements, as shown in
Fig.~\ref{fig:intro}(a). These difficulties become more pronounced as
object counts and spatial relationships grow more complex. Meanwhile,
compositional scene data remain limited in quantity and
diversity~\cite{dai2017scannet,fu20213dfront}, providing limited
supervision for jointly learning global arrangements and fine-grained
object geometry. This motivates a division of learning responsibilities:
scene-level supervision establishes a coherent scaffold, while a
pretrained object generator supplies detailed geometry and appearance
priors. An explicit spatial interface between these two capabilities
allows scene organization to guide detailed object synthesis.

In contrast, object-centric generators are trained on abundant
collections of clean and normalized shapes, such as Objaverse-XL
~\cite{deitke2023objaverse,deitke2024objaversexl}, and therefore learn
strong geometry and appearance priors. However, these priors are
designed for isolated objects rather than multiple instances embedded
in a shared scene coordinate system. Directly extending an
object-centric generator to synthesize an entire scene in a single pass
requires its representation capacity and spatial resolution to be
distributed across all objects and their arrangements. A resolution
sufficient for a single object therefore becomes too coarse at the
scene scale, often resulting in incomplete geometry, distorted shapes,
and incorrect orientations, as illustrated in
Fig.~\ref{fig:intro}(b). In addition, the object
representation cannot be directly aligned with the scale
and orientation of each scene instance.

These observations motivate a different route for transferring strong
object-level generative priors to scene generation. We use a shared
spatial scaffold as the intermediate representation connecting
scene-level structure with object-level generation. Such a
representation should establish globally coordinated layouts in
a shared scene space, while remaining decomposable into individual
instances that can be processed by a pre-trained object generator.
It should further retain the anchor-defined spatial reference of
each instance while providing sufficient object-scale resolution for
high-fidelity geometry and appearance synthesis.

To bridge this gap, we propose Bri3D, an image-to-compositional 3D scene generation framework built upon a scene scaffold that serves as the intermediate representation between scene-level structure and object-level generation.
Bri3D therefore uses scene anchors to connect scene-level layout
learning with pretrained object-level detail synthesis, reducing the
need to learn both capabilities jointly from scene data.
Bri3D first constructs a unified, instance-aware scene scaffold. This
scaffold provides a globally coordinated structural anchor that jointly
captures coarse object geometry and spatial arrangements.
Specifically, the \textit{Scene Scaffold Generator} predicts a
color-coded, instance-aware voxel scaffold from an RGB image and its
corresponding segmentation result. Since all foreground objects are generated
within a shared scene representation, their placements and coarse
geometry are jointly determined under a common scene context. The
predicted color labels separate the scaffold into coarse object voxel
clusters, each serving as a \textit{scene anchor} for object-scale
refinement. The \textit{Anchored Voxel Refiner} encodes each clustered voxel into a
structural latent and injects Gaussian noise to construct a
\textit{scene-anchored latent initialization}. Conditioned on the
corresponding image evidence, it denoises the latent into a
pose-preserving, prior-compatible voxel support. This support is then
processed by the frozen detail-synthesis module of a pre-trained object
generator to synthesize a high-fidelity textured 3D object. The
generated objects are finally transformed back to their scaffold-derived
spatial configurations and composed into the complete scene.

In summary, our contributions are threefold:

\begin{itemize}
    \item We propose Bri3D, an image-to-compositional 3D scene
    generation framework that bridges coordinated scene-level modeling
    and detailed object-level synthesis through instance-aware voxel
    scene anchors.

    \item We introduce a \textit{Scene Scaffold Generator} that predicts
    a unified, instance-aware voxel scaffold encoding globally
    coordinated object arrangements and coarse geometry.

    \item We develop an \textit{Anchored Voxel Refiner} that converts
    clustered scene anchors into pose-preserving, prior-compatible voxel
    supports for a frozen object-level detail-synthesis module.
\end{itemize}

\section{Related Work}

\subsection{3D Object Generation}

Recent 3D object generators achieve high geometric fidelity by learning
from large-scale collections of isolated shapes. Existing methods mainly
use point-, voxel-, or mesh-based representations. Point-based methods,
such as 3DShape2VecSet~\cite{zhang20233dshape2vecset}, encode unordered
point sets into compact latents, while diffusion and rectified-flow
models further improve generation quality and scalability
~\cite{zhang2024clay,li2025triposg}. Mesh-based methods model
surface connectivity and produce high-quality geometry, but
autoregressive decoding often incurs substantial inference cost
~\cite{siddiqui2024meshgpt}. Voxel-based methods provide explicit spatial structure and are
well suited for image-conditioned generation. TRELLIS
~\cite{xiang2025structured,xiang2026native} introduces Sparse Structured Latents that
combine sparse voxel supports with per-voxel features and support
decoding into multiple 3D representations. Subsequent methods further
improve geometric fidelity and efficiency
~\cite{ye2025hi3dgen,wu2025direct3d}. However, these models are trained
on isolated, normalized shapes and lack the scene-level reasoning
required for object placement, scale, orientation, and spatial
coordination. Directly extending them to full-scene generation therefore
often degrades structural quality.

\subsection{Compositional 3D Scene Reconstruction}

A common strategy for single-image scene reconstruction is to generate
or retrieve individual objects and place them in a shared 3D coordinate
system. Earlier methods rely on CAD retrieval and alignment
~\cite{kuo2020mask2cad,gumeli2022roca}, while recent approaches use scene
graphs, geometric cues, or vision-language models to infer object
identities and spatial configurations
~\cite{gao2024graphdreamer,yao2025cast}. Although these pipelines can
reuse high-quality assets, they depend heavily on monocular depth, pose,
and scale estimation. Errors propagate through registration and
assembly, producing floating objects, intersections, or inconsistent
layouts. Independently generated objects may also be geometrically
incompatible, limiting global scene coherence. SceneDreamer
~\cite{chen2023scenedreamer} learns unbounded 3D landscape generation
from 2D image collections through adversarial training. Our focus is
single-image reconstruction of compositional scenes, where object
geometry and relative placement must jointly explain the observation.

\subsection{Object-to-Scene Generation}

Recent methods extend object priors to scene reconstruction through
different forms of spatial reasoning. MIDI~\cite{huang2025midi}
introduces multi-instance attention to coordinate object generation.
SceneGen~\cite{meng2025scenegen} adds global attention and position
prediction to a frozen object generator. SAM3D
~\cite{sam3dteam2025sam3d3dfyimages} predicts object shape and pose
from a target mask and image context. Other approaches use hierarchical
or compositional representations to model scene layout and local
geometry~\cite{meng2025lt3sd,lin2025partcrafter}.

Structured or staged processing is also explored across 3D vision tasks~\cite{hou2024detailed,hou2026reference,du2026hierarchical,zhang2026citystar,li2026holorecon}.

These methods demonstrate the value of object priors for scene synthesis.
Bri3D connects these priors to scene organization through an explicit,
shared voxel scaffold. The scaffold jointly represents coarse geometry,
instance identity, and placement, and is decomposed into spatially
anchored object conditions. Refinement then adapts each coarse anchor to
the detail generator while retaining its scene-defined coordinate
reference. This interface assigns global spatial coordination to the
scene model and detailed synthesis to the object prior, allowing the
two levels to operate at their respective spatial resolutions.

\begin{figure*}[t]
    \centering
    \includegraphics[width=\textwidth]{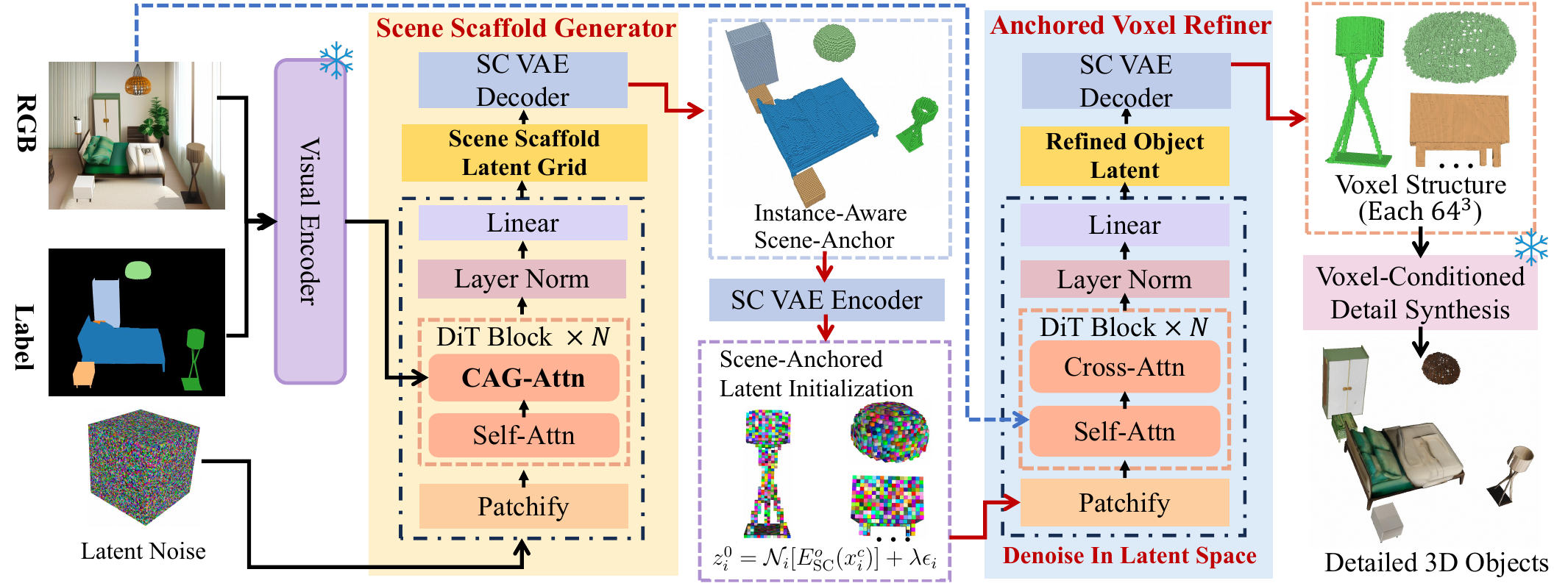}
    \caption{\textbf{Overview of Bri3D.} The Scene Scaffold Generator and Anchored Voxel Refiner replace
    the voxel-generation stage of a two-stage object generator, while
    its detail-synthesis stage remains frozen.
    SC-VAE denotes the Structure-Color VAE, and snowflakes indicate
    frozen modules. The scaffold is generated in latent space and decoded
    at $128^3$; each object is refined to a $64^3$ voxel support.
    The anchor initialization follows Eq.~\ref{eq:scene_anchor}, including
    object-wise latent normalization.
}
    \label{fig:pipeline}
\end{figure*}

\Needspace{6\baselineskip}
\section{Method}
\label{sec:method}

\subsection{Overview}
\label{sec:overview}

Bri3D transfers the strong geometry and appearance priors of a
pre-trained single-object generator to compositional scene generation
through a unified, instance-aware scene scaffold. Rather than generating
objects independently and relying primarily on inter-instance
interactions, Bri3D first establishes their coarse geometry and spatial
arrangements within a shared scene representation. It contains two
trainable modules: a \textit{Scene Scaffold Generator}
$\mathcal{G}_{S}$, which predicts a color-coded voxel scaffold for all
foreground objects, and an \textit{Anchored Voxel Refiner}
$\mathcal{R}_{A}$, which converts each scaffold-derived object cluster
into a high-resolution voxel support compatible with the pre-trained
object generator.
The key to our design is to use voxel structure as the interface
between scene-level spatial organization and object-level detail
synthesis. The scene scaffold establishes coarse object geometry
and spatial relationships in a shared coordinate system, providing
structural anchors for subsequent refinement. Guided by these anchors
and object-specific image evidence, the Anchored Voxel
Refiner recovers finer object geometry within the spatial reference
established by the scaffold. The refined voxel supports
then condition a frozen detail-synthesis module from a pre-trained
object generator~\cite{xiang2025structured} to produce detailed
geometry and appearance. Finally, the generated objects are mapped
back into the shared scene using their scaffold-derived spatial
transformations. This design allows Bri3D to learn scene organization
and anchored geometry refinement while reusing pre-trained
object-level priors for detail synthesis.

\subsection{Scene Scaffold Generator}
\label{sec:scaffold_generator}

The Scene Scaffold Generator $\mathcal{G}_{S}$ predicts the coarse voxel
geometry and instance organization of all foreground objects. Its
training target is a color-coded occupancy volume
$\mathbf{x}_{s}$ at a resolution of $128^3$. Empty voxels represent
free space, while each occupied voxel carries the color label of its
corresponding object instance. Distinct instance colors allow adjacent
objects, including instances from the same category, to remain
separable within the shared scene representation. At voxel $j$, we use
the four-channel representation $\mathbf{v}_j=[o_j,r_j,g_j,b_j]^\top$,
where $o_j\in\{0,1\}$ denotes occupancy and
$(r_j,g_j,b_j)\in[0,1]^3$ encodes instance identity, not object texture.
Thus, the scaffold separates instance labeling from the appearance
synthesized by the frozen object generator.

\textbf{Structure-color latent representation: }We employ a scene-scale Structure-Color VAE (SC-VAE) to encode the
color-coded scaffold into a spatial latent grid:
\begin{equation}
    \mathbf{Z}_{s}
    =
    E_{\mathrm{SC}}^{s}
    \left(
    \mathbf{x}_{s}
    \right),
    \qquad
    \hat{\mathbf{x}}_{s}
    =
    D_{\mathrm{SC}}^{s}
    \left(
    \mathbf{Z}_{s}
    \right).
\label{eq:scene_scvae}
\end{equation}
Unlike a single global latent code, $\mathbf{Z}_{s}$ retains a regular
3D organization in which each token corresponds to a local scene
region. $\mathbf{Z}_{s}\in
\mathbb{R}^{h_s\times w_s\times d_s\times c_s}$, it therefore preserves local object structure while providing
global spatial context for scene generation.

We adopt the same SC-VAE design at the scene and object scales, with
superscripts $s$ and $o$ denoting their corresponding
resolution-specific models.

\textbf{Color-aware conditional generation: }A rectified-flow DiT $\mathcal{G}_S$ generates the scene scaffold latent
  $\mathbf{Z}_{s}$ from Gaussian noise of the same latent-grid dimensions.
  Decoding yields a $128^3$ scene scaffold. The RGB image
  $I_{\mathrm{rgb}}$ and its color-coded instance label map
  $I_{\mathrm{lab}}$ are encoded by a shared
  DINOv2~\cite{oquab2023dinov2} backbone, producing appearance and
  instance-layout features.

  Instead of directly concatenating these conditioning modalities, we
  introduce Color-Attention Gating (CAG) in each cross-attention block.
  For the $k$-th latent query $\mathbf{q}_{k}$ and conditioning modality
  $m\in\{\mathrm{rgb},\mathrm{lab}\}$, the corresponding gating
  coefficient is
  \begin{equation}
      g_{m}^{k}
      =
      \sigma
      \left(
      \mathrm{MLP}_{m}
      \left(
      \mathbf{q}_{k}
      \right)
      \right),
  \label{eq:cag_gate}
  \end{equation}
  and the conditioning features are fused as
  \begin{equation}
      \mathbf{q}_{k}^{\prime}
      =
      \mathbf{q}_{k}
      +
      \sum_{m\in\{\mathrm{rgb},\mathrm{lab}\}}
      g_{m}^{k}
      \,
      \mathrm{Attn}_{m}
      \left(
      \mathbf{q}_{k},
      \mathbf{F}_{m}
      \right),
  \label{eq:cag_fusion}
  \end{equation}
  where $\mathbf{F}_{m}$ denotes the feature tokens of modality $m$.
  CAG allows each spatial region to adaptively select appearance or
  instance-layout evidence according to color cues. Since the instance
  colors in $I_{\mathrm{lab}}$ are assigned arbitrarily, CAG is designed
  to be robust to the specific color assignment and to color
  perturbations; a robustness analysis under imperfect instance label
  maps is provided in Section~\ref{sec:ablation}. The generated latent grid is finally decoded by
  $D_{\mathrm{SC}}^{s}$ into the instance-aware scene scaffold
  $\hat{\mathbf{x}}_{s}$.

\subsection{Anchored Voxel Refiner}
\label{sec:refiner}

The Anchored Voxel Refiner $\mathcal{R}_{A}$ converts each coarse object
extracted from the scene scaffold into a high-resolution voxel support
compatible with the pre-trained object generator. It takes a clustered
coarse voxel and its corresponding image evidence as input, using the
voxel structure as a scene anchor for object-scale refinement.

\textbf{Instance extraction: }We apply DBSCAN to occupied voxels using
both spatial positions and predicted instance colors. Spatial proximity
and color similarity jointly determine the clusters, accommodating
continuous color predictions without requiring exact RGB equality.
This helps distinguish touching instances that geometry-only connected
component analysis would merge. Each cluster retains its coordinates
in the shared scene and becomes an object anchor for normalization
and refinement. This inference-time clustering differs from the K-means
partitioning used to construct single-object training labels.

\textbf{Anchor-defined spatial reference: }We normalize each scene
anchor using its center $\mathbf{c}_i$ and isotropic scale $s_i>0$,
refine its geometry in the resulting local coordinate frame, and map
the output back using the same inverse transformation:
\begin{equation}
    \mathbf{u}=\frac{\mathbf{x}-\mathbf{c}_i}{s_i},
    \qquad
    \hat{\mathbf{x}}=s_i\hat{\mathbf{u}}+\mathbf{c}_i.
    \label{eq:anchor_coordinates}
\end{equation}
Here, $\mathbf{x}$ is a point in scene coordinates, $\mathbf{u}$ is its
local coordinate, and $\hat{\mathbf{u}}$ is an output point after
refinement and detail synthesis. We retain the original
$(\mathbf{c}_i,s_i)$ throughout the pipeline; the output is not
re-normalized using its predicted bounding box. Normalization applies
only translation and isotropic scaling, so the anchor's orientation
remains encoded in the local voxels without a separate rotation estimate.
Here, \emph{pose preservation} refers to retaining the anchor-defined
spatial reference without additional pose estimation or registration.
Object geometry is refined within this fixed spatial reference.
We denote the normalized coarse voxel structure by $\mathbf{x}_i^c$.

\textbf{Scene-anchored latent initialization: } The coarse voxel is encoded by the object-scale SC-VAE, and Gaussian
noise is injected into its structural latent:
\begin{equation}
    \mathbf{z}_{i}^{0}
    =
    \mathcal{N}_i\!\left[E_{\mathrm{SC}}^{o}
    (\mathbf{x}_{i}^{c})\right]
    +
    \lambda
    \boldsymbol{\epsilon}_{i},
    \qquad
    \boldsymbol{\epsilon}_{i}
    \sim
    \mathcal{N}
    \left(
    \mathbf{0},
    \mathbf{I}
    \right).
\label{eq:scene_anchor}
\end{equation}
Here, $\mathcal{N}_i$ uses the coarse anchor latent's per-channel
mean and standard deviation for both condition and target normalization;
the same statistics are saved for inverse normalization before decoding
(Appendix~\ref{app:latent_norm}). We refer to $\mathbf{z}_{i}^{0}$ as the
\textit{scene-anchored latent initialization}. Its encoded component
preserves the scaffold-derived pose and coarse geometry, while the noise
provides the flexibility required to recover missing and unobserved
structures.

\textbf{Multi-resolution training conditions: }We construct coarse
conditions from ground-truth (GT) object voxel grids rather than requiring
predicted scene anchors for every training example. Each GT object is
voxelized at resolutions $r^3$, with $r\in\{4,8,16,32\}$, and the coarse
grid is restored to the fixed $64^3$ input grid before entering the
object-scale encoder:
\begin{equation*}
    \mathbf{x}_{i,r}^{c}=\mathcal{U}_{64}
    \bigl(\mathcal{V}_{r}(\mathbf{x}_i^{\mathrm{GT}})\bigr),
    \qquad
    \mathbf{z}_{i,r}^{\mathrm{cond}}=\mathcal{N}_i\!\left[E_{\mathrm{SC}}^{o}
    (\mathbf{x}_{i,r}^{c})\right].
\end{equation*}
Here, $\mathcal{V}_{r}$ denotes coarse voxelization and
$\mathcal{U}_{64}$ restores the grid resolution while retaining the
coarse geometric information. Gaussian noise is added to this encoded
condition as in Eq.~\ref{eq:scene_anchor}; the full-resolution GT grid
provides the clean target latent. These four levels expose the refiner
to varying amounts of structural detail during training.

\textbf{Image-conditioned latent refinement: }Starting from $\mathbf{z}_{i}^{0}$, a Geometry Flow Transformer
progressively recovers a refined object latent. The refiner is
conditioned on an image feature $\mathbf{f}_{i}$ extracted from the
corresponding object region.
The refinement trajectory is formulated as
\begin{equation}
    \frac{d\mathbf{z}_{i}}{dt}
    =
    \mathbf{v}_{\theta}
    \left(
    \mathbf{z}_{i},
    t;
    \mathbf{f}_{i}
    \right),
    \qquad
    \mathbf{z}_{i}(0)
    =
    \mathbf{z}_{i}^{0}.
\label{eq:refiner_flow}
\end{equation}
The initialization carries the coarse anchor geometry, while image
conditioning guides the recovery of object-specific geometry.

After inverse normalization, the object-scale SC-VAE decodes
$\mathcal{N}_i^{-1}(\hat{\mathbf{z}}_i)$ into a four-channel occupancy--RGB volume
$\hat{\mathbf{x}}_i^o\in[0,1]^{64\times64\times64\times4}$.
We explicitly select its occupancy channel:
\begin{equation}
    \hat{\mathbf{V}}_{i}
    =
    \pi_{\mathrm{occ}}\!\left[
    D_{\mathrm{SC}}^{o}(\mathcal{N}_i^{-1}(\hat{\mathbf{z}}_{i}))
    \right],
    \qquad
    \hat{\mathbf{V}}_{i}
    \in
    [0,1]^{64\times64\times64}.
\label{eq:object_decode}
\end{equation}
Here, $\pi_{\mathrm{occ}}$ selects the occupancy channel and discards
the three instance-color channels. We threshold this field to obtain
the binary support $\mathbf{B}_i(\mathbf{p})=\mathbf{1}
[\hat{\mathbf{V}}_i(\mathbf{p})>\tau_{\mathrm{occ}}]$, where
$\tau_{\mathrm{occ}}$ is the occupancy threshold. Only occupied
locations $\{\mathbf{p}:\mathbf{B}_i(\mathbf{p})=1\}$ are passed to
the object generator as its sparse structural support.

\textbf{Voxel-conditioned detail synthesis:} The thresholded voxel support $\mathbf{B}_{i}$ is passed to the frozen
detail-synthesis module of the pre-trained object generator \cite{xiang2025structured}. Conditioned
on the occupied voxel and the corresponding image,
this module generates detailed geometry and appearance to synthesize a
high-fidelity textured 3D object. The generated objects are transformed
back using the same saved $(\mathbf{c}_i,s_i)$ in
Eq.~\ref{eq:anchor_coordinates} and composed into the final 3D scene,
without an additional registration step.

\subsection{Training Objectives}
\label{sec:loss}

\textbf{SC-VAE objective: }The scene-scale and object-scale SC-VAEs are trained to reconstruct
structure-color voxel representations. Given an input volume
$\mathbf{x}$ and its reconstruction $\hat{\mathbf{x}}$, the objective
is
\begin{equation}
\begin{split}
    \mathcal{L}_{\mathrm{SC}}
    =
    &\;
    \lambda_{\mathrm{rec}}
    \left\|
    \mathbf{x}
    -
    \hat{\mathbf{x}}
    \right\|_{1}
    +
    \beta
    \,
    \mathrm{KL}
    \left(
    q(\mathbf{Z}\mid\mathbf{x})
    \,\Vert\,
    p(\mathbf{Z})
    \right)
    \\
    &+
    \lambda_{\mathrm{render}}
    \mathcal{L}_{\mathrm{render}}.
\end{split}
\label{eq:scvae_loss}
\end{equation}
The reconstruction term preserves voxel occupancy and color
information, while the KL term regularizes the spatial latent
distribution.

To better preserve projected object boundaries and spatial
arrangements, ground-truth and reconstructed instance-color volumes
are rendered from the same camera to obtain label images $I_{\mathrm{gt}}$
and $I_{\mathrm{pred}}$. The rendering loss is defined as
\begin{equation}
\mathcal{L}_{\mathrm{pixel}}
=
\left\|I_{\mathrm{gt}}-I_{\mathrm{pred}}\right\|_{1}
+
0.2\left[
1-\mathrm{SSIM}
\left(I_{\mathrm{gt}},I_{\mathrm{pred}}\right)
\right].
\label{eq:pixel_loss}
\end{equation}
\begin{equation}
\mathcal{L}_{\mathrm{render}}
=
\mathcal{L}_{\mathrm{pixel}}
+
0.2\,\mathrm{LPIPS}
\left(I_{\mathrm{gt}},I_{\mathrm{pred}}\right).
\label{eq:render_loss}
\end{equation}
This supervision improves view-consistent object separation and
scene-level spatial organization.

\textbf{Rectified flow matching: }Both the Scene Scaffold Generator and Anchored Voxel Refiner are
trained using rectified flow matching. Given an initial latent
$\mathbf{z}_{0}$ and a target latent $\mathbf{z}_{1}$, we define the
linear interpolation
\begin{equation*}
    \mathbf{z}(t)
    =
    (1-t)\mathbf{z}_{0}
    +
    t\mathbf{z}_{1},
    \qquad
    t\sim\mathcal{U}[0,1],
\end{equation*}
and optimize
\begin{equation}
    \mathcal{L}_{\mathrm{flow}}
    =
    \mathbb{E}_{t}
    \left[
    \left\|
    \mathbf{v}_{\theta}
    \left(
    \mathbf{z}(t),
    t;
    \mathbf{f}
    \right)
    -
    \left(
    \mathbf{z}_{1}
    -
    \mathbf{z}_{0}
    \right)
    \right\|_{2}^{2}
    \right].
\label{eq:flow_loss}
\end{equation}

Here, $\mathbf{f}$ denotes the RGB/label conditioning features for scene
scaffold generation and the object image feature $\mathbf{f}_i$ for
refinement. For the Scene Scaffold Generator, $\mathbf{z}_{0}$ is Gaussian noise
and $\mathbf{z}_{1}$ is the SC-VAE latent of the ground-truth scene
scaffold. For the Anchored Voxel Refiner, $\mathbf{z}_{0}$ is the
scene-anchored initialization defined in
Eq.~\ref{eq:scene_anchor}, while $\mathbf{z}_{1}$ is the object-scale
SC-VAE latent of the corresponding ground-truth voxel structure,
normalized using the coarse anchor's saved statistics.

\section{Experiments}

\subsection{Experimental Setup}\label{sec:experimental_setup}

\textbf{Implementation details.} Bri3D trains a Scene Scaffold Generator and an Anchored Voxel Refiner using Objaverse-XL~\cite{deitke2024objaversexl} and 3D-FRONT~\cite{fu20213dfront}. Each 3D-FRONT instance is assigned a unique color, while Objaverse-XL objects are partitioned into 1--6 K-means clusters with distinct colors to construct structure-color supervision. An SC-VAE encodes and reconstructs these color-coded voxel structures with rasterization-based rendering losses, providing latents for both scaffold generation and object refinement. The generator predicts a $128^3$ instance-aware scaffold, and the refiner converts each object cluster into a prior-compatible $64^3$ voxel structure. Random occluders improve robustness to partial observations. Bri3D reconstructs foreground objects and their layouts, excluding floors, walls, and ceilings. Scene-level training follows MIDI's 3D-FRONT split~\cite{huang2025midi},
drawn from its cleaned corpus of approximately 15,000 scenes.
Separately, the object-level data construction uses 500,000 single-object
instances to provide coarse conditions and GT targets for refinement.
Both the Scene Scaffold Generator and Anchored Voxel Refiner are trained
for 10 epochs on 16 NVIDIA H20 GPUs, using AdamW with a learning rate
of $1\times10^{-4}$ and a per-GPU batch size of 4.

\textbf{Evaluation metrics.}
We evaluate geometry, layout, and visual quality following existing 3D scene generation methods~\cite{huang2025midi,meng2025scenegen}. Geometric fidelity is measured using Chamfer Distance (CD) and F-Score, computed at both the scene and object levels. We use CD-S/CD-O and F-S/F-O for scene/object metrics throughout; all reported CD values are scaled by $100$, and F-S/F-O are percentages. Spatial layout accuracy is assessed using the Intersection over Union
(BBox IoU) between corresponding axis-aligned bounding boxes in the
generated and ground-truth scenes. For Bri3D, BBox IoU evaluates the
scaffold-derived anchor boxes, which are retained through refinement
and detail synthesis. Collision rate is the percentage of object pairs
whose voxelized occupancies overlap in shared scene coordinates. For
Bri3D, scaffold instances are extracted and voxelized separately before
refinement; collisions are measured between these instance volumes. For visual quality, we compare rendered views of the generated scenes with corresponding ground-truth images to evaluate texture realism and perceptual fidelity.
Multiple perceptual metrics are reported, including PSNR, SSIM, LPIPS~\cite{zhang2018unreasonable}, FID~\cite{heusel2017gans}, CLIP similarity~\cite{radford2021clip}, and DINOv2 similarity~\cite{oquab2023dinov2}.

\textbf{Baselines.}
We compare Bri3D against representative 3D scene generation methods: PartCrafter~\cite{lin2025partcrafter}, DepR~\cite{zhao2025depr}, Gen3DSR~\cite{ardelean2025gen3dsr}, MIDI~\cite{huang2025midi}, SceneGen~\cite{meng2025scenegen}, and SAM3D ~\cite{sam3dteam2025sam3d3dfyimages}.
Since PartCrafter and DepR do not support texture rendering, we exclude them from visual-quality evaluation.
In-domain quantitative evaluation uses MIDI's released 3D-FRONT~\cite{fu20213dfront,huang2025midi} test split of 1,000 images, with the corresponding 3D scenes and instance masks. Consistent with the evaluation protocol of MIDI~\cite{huang2025midi}, we use the instance segmentation masks provided by the dataset, rather than masks predicted by an off-the-shelf segmentor, as the segmentation input for standard quantitative comparisons. The robustness of Bri3D to erroneous segmentation maps is further analyzed in our ablation study (Fig.~\ref{fig:more_scenes}). For cross-dataset evaluation, we manually select 300 scenes: 150 from
HSSD~\cite{khanna2024habitat} and 150 independently collected from
the BlendSwap platform~\cite{blendswap}. We render multiple input views
per scene ourselves, pairing each view with its GT 3D scene; each
reconstruction uses one input image, so 300 denotes the number of scenes,
not the number of input views. ScanNet~\cite{dai2017scannet}, which lacks
complete ground truth, is used for qualitative evaluation.

\begin{table*}[!t]
\caption{\textbf{Quantitative comparisons on the 3D-FRONT~\cite{fu20213dfront} test set.}
Geometric metrics: scene/object-level Chamfer Distance (CD-S/CD-O, $\times100$), F-Score (F-S/F-O, \%), and BBox IoU ($[0,1]$).
Visual metrics: PSNR, SSIM, LPIPS, FID, CLIP-S, DINO-S.}
\label{tab:scene_metrics}
\centering
\small

\setlength{\tabcolsep}{3.6pt}
\begin{tabular}{l ccccc | cccccc}
\hline
\multirow{2}{*}{Method}
& \multicolumn{5}{c|}{\textbf{Geometry}}
& \multicolumn{6}{c}{\textbf{Visual Quality}} \\
\cline{2-6} \cline{7-12}
& CD-S$\downarrow$ & CD-O$\downarrow$ & F-S$\uparrow$ & F-O$\uparrow$ & BBox IoU$\uparrow$ %
& PSNR$\uparrow$ & SSIM$\uparrow$ & LPIPS$\downarrow$ & FID$\downarrow$ & CLIP-S$\uparrow$ & DINO-S$\uparrow$ \\
\hline
DepR
& 9.07 & 18.81 & 78.91 & 56.15 & 0.3943
& ---  & ---  & ---  & ---  & ---  & --- \\
Gen3DSR
& 8.21 & 17.18 & 79.07 & 60.41 & 0.4456
& 16.30 & 0.8639 & 0.1621 & 19.96 & 0.9009 & 0.8139\\
PartCrafter
& 7.05 & ---    & 85.66 & ---    & ---
& ---  & ---  & ---  & ---  & ---  & ---  \\
SceneGen
& 6.11 & 7.07 & 87.72 & 87.73 & 0.5382
& 16.50 & 0.8723 & 0.1552 & 19.80 & 0.9005 & 0.8201\\
MIDI
& 2.14 & 2.81 & 92.20 & 89.57 & 0.6238
& 16.97 & 0.8756 & 0.1522 & 19.78 & 0.9315 & 0.8247 \\
SAM3D
& 2.11 & 2.76 & 92.34 & 89.72 & 0.6271
& 17.02 & 0.8791 & 0.1496 & 19.71 & 0.9328 & 0.8295 \\
Ours
& \textbf{0.87} & \textbf{1.72} & \textbf{96.12} & \textbf{95.33} & \textbf{0.7540}
& \textbf{17.20} & \textbf{0.9020} & \textbf{0.1312} & \textbf{19.27} & \textbf{0.9343} & \textbf{0.8366} \\
\hline
\end{tabular}

\end{table*}
\begin{table}[!t]
\caption{
\textbf{Cross-dataset generalization results.}
We evaluate scene- and object-level geometry, layout accuracy, and visual similarity on 300 scenes paired with rendered images and ground-truth 3D scenes. CD-S and CD-O are scaled by $100$; F-S and F-O are F-Score percentages; BBox IoU is on the $[0,1]$ scale. The best results are in bold; the second-best results (SAM3D in all reported columns) are underlined.
}
\label{tab:new_dataset_generalization}
\centering
\scriptsize
\setlength{\tabcolsep}{2.6pt}
\renewcommand{\arraystretch}{1.05}

\resizebox{\columnwidth}{!}{
\begin{tabular}{lccccccc}
\toprule
\textbf{Method}
& \textbf{CD-S}$\downarrow$
& \textbf{F-S}$\uparrow$
& \textbf{CD-O}$\downarrow$
& \textbf{F-O}$\uparrow$
& \textbf{BBox IoU}$\uparrow$
& \textbf{LPIPS}$\downarrow$
& \textbf{DINO-S}$\uparrow$ \\
\midrule

DepR
& 14.20
& 64.80
& 13.85
& 66.10
& 0.2420
& -
& - \\

Gen3DSR
& 12.70
& 68.90
& 12.35
& 69.70
& 0.2785
& 0.3328
& 0.7218 \\

PartCrafter
& 11.30
& 74.60
& ---
& ---
& ---
& -
& - \\

SceneGen
& 12.10
& 71.20
& 11.56
& 72.80
& 0.2864
& 0.3187
& 0.7382 \\

MIDI
& 6.20
& 80.50
& 5.74
& 82.60
& 0.4315
& 0.2635
& 0.7756 \\

SAM3D
& \underline{3.85}
& \underline{87.10}
& \underline{3.48}
& \underline{89.30}
& \underline{0.5628}
& \underline{0.2218}
& \underline{0.8021} \\

Ours
& \textbf{3.42}
& \textbf{88.20}
& \textbf{3.06}
& \textbf{90.60}
& \textbf{0.5946}
& \textbf{0.2076}
& \textbf{0.8104} \\

\bottomrule
\end{tabular}
}

\end{table}
\begin{figure*}[!t]
    \centering
    \includegraphics[width=0.95\textwidth]{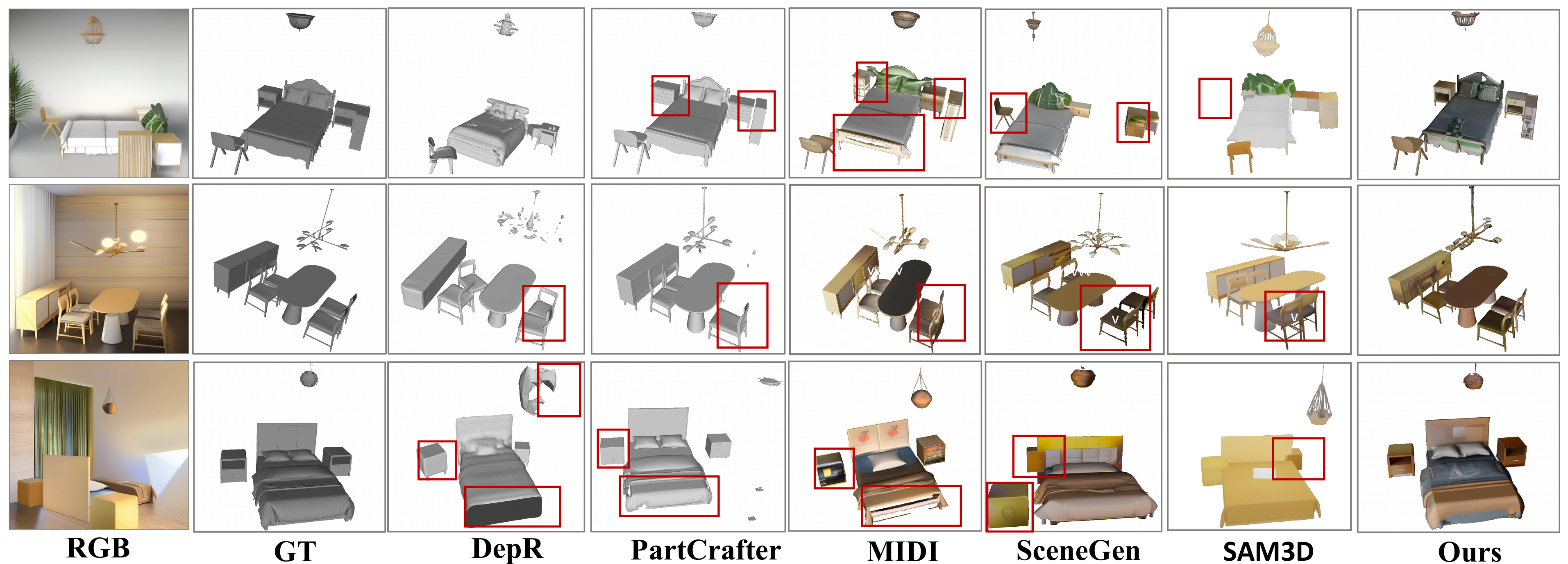}
    \caption{\textbf{Qualitative results on the 3D-FRONT~\cite{fu20213dfront} test set.}
Our method generates more plausible layouts and higher-fidelity geometry, consistent with the quantitative improvements. Red boxes indicate inferior baseline results.
}
    \label{fig:result}
\end{figure*}
\begin{figure*}[!t]
    \centering
    \includegraphics[width=0.9\linewidth]{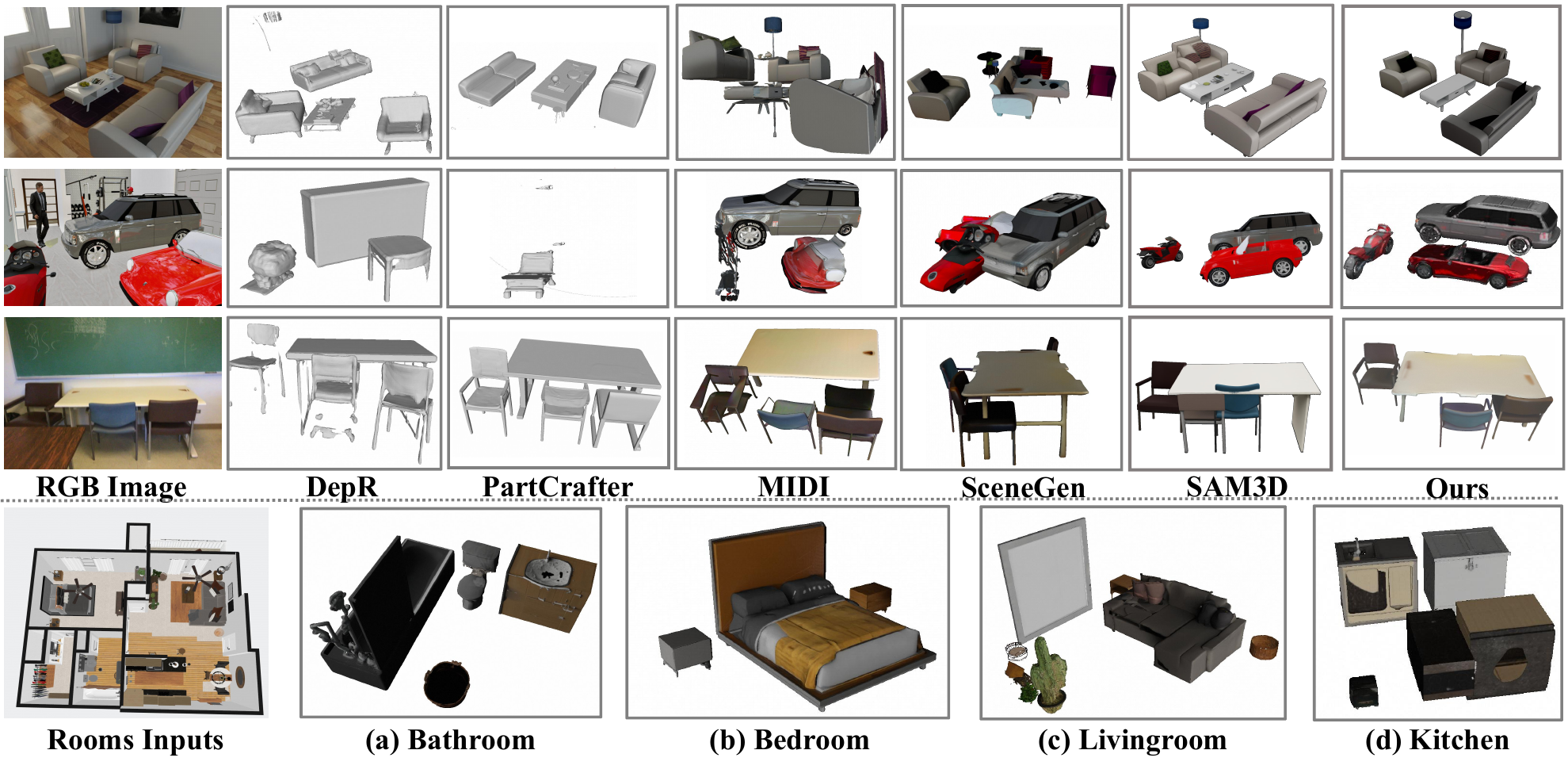}
    \caption{\textbf{Generalization to diverse room scenes.}
    The upper three rows compare reconstructions of a living room, a garage,
    and a classroom, with the input image followed by the results of DepR,
    PartCrafter, MIDI, SceneGen, SAM3D, and Bri3D. These examples cover
    different object categories, scales, and degrees of occlusion. The lower
    row shows a multi-room overview and reconstructions of its bathroom,
    bedroom, living room, and kitchen, each generated from a separate
    room-level input. Together, the examples illustrate
    object reconstruction across varied scene configurations and room types.}
    \label{fig:figure4}
\end{figure*}
\begin{table}[!t]
\caption{
    \textbf{Comparison of layout accuracy, inference time, and observed
    object collisions.}
    BBox IoU is in $[0,1]$; collision rate is the percentage of overlapping object pairs. $^{*}$Ours reports scaffold-stage anchor BBox IoU and collisions after separately voxelizing extracted instances. Times outside parentheses are for scaffold prediction; parentheses and baseline times report full generation.
    }
\label{tab:ObjectCountIOUTime}
    \footnotesize
    \centering

    \resizebox{\linewidth}{!}{%
    \begin{tabular}{l|cc|cc|c}
        \toprule
        \multirow{2}{*}{Model}
        & \multicolumn{2}{c|}{1--10 Objects}
        & \multicolumn{2}{c|}{11--20 Objects}
        & \multirow{2}{*}{Collision Rate (\%)$\downarrow$} \\
        & BBox IoU & Test Time & BBox IoU & Test Time & \\
        \midrule
        DepR
        & 0.4032 & 15.3s
        & 0.3008 & 20.8s
        & 22\\

        MIDI
        & 0.6349 & 40.4s
        & 0.4193 & 72.9s
        & 14\\

        SceneGen
        & 0.5892 & 135.0s
        & 0.4462 & 230.2s
        & 38\\

        SAM3D
        & 0.6568 & 64.1s
        & 0.4609 & 179.6s
        & 34 \\

        Ours$^{*}$
        & \textbf{0.7980} & 5.1s (42.6s)
& \textbf{0.6732} & 5.2s (78.4s)
        & \textbf{4} \\
        \bottomrule
    \end{tabular}%
    }

\end{table}

\subsection{Scene Generation}
As shown in Table~\ref{tab:scene_metrics}, our method outperforms
state-of-the-art approaches across both scene-level and object-level
metrics, achieving lower Chamfer Distance (CD-S: 0.87, CD-O: 1.72) and
higher BBox IoU (0.7540). These improvements demonstrate better global
layout fidelity and fine-grained object geometry. Qualitative comparisons
in Fig.~\ref{fig:result} further show that Bri3D produces coherent object
arrangements while reducing floating objects, unnatural occlusions, and
geometric distortions. The cross-dataset results in
Table~\ref{tab:new_dataset_generalization} further demonstrate strong
generalization to challenging unseen scenes. Bri3D consistently achieves
the best scene- and object-level geometry, layout accuracy, and visual
similarity, while SAM3D obtains the second-best results. These results
indicate that the scene-anchor representation effectively transfers
object-level generative priors beyond the training distribution. As shown in Fig.~\ref{fig:figure4}, Bri3D also generalizes to diverse room
layouts with heavy occlusions and incomplete observations, producing
high-fidelity reconstructions under challenging viewing conditions.
The room-level examples span bathrooms, bedrooms, living rooms, and
kitchens, showing coherent object arrangements across varied room
functions and furniture configurations.

\begin{figure}[!t]
    \centering
    \includegraphics[width=0.95\linewidth]{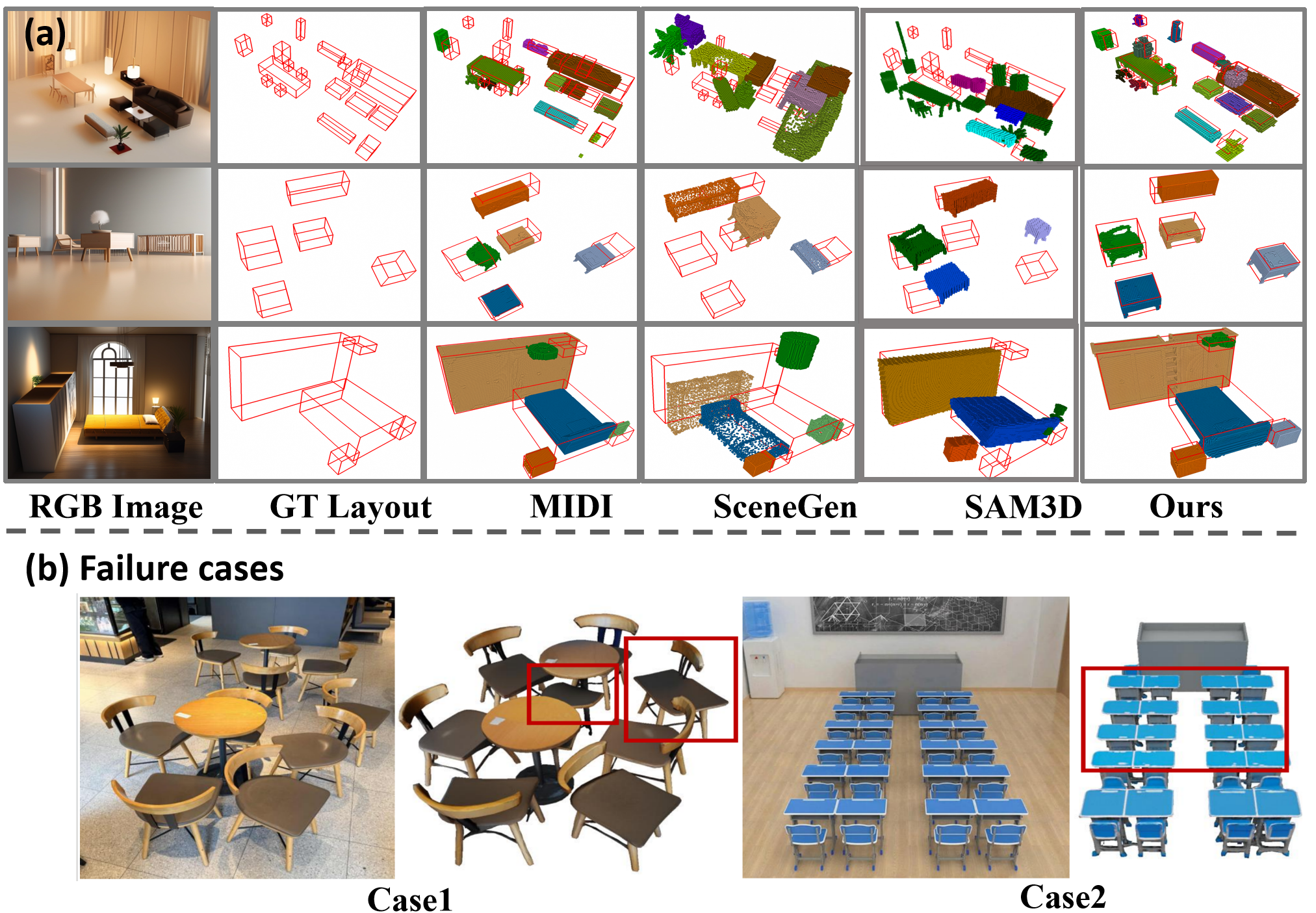}
    \caption{
        \textbf{Comparison of 3D scene layouts and failure cases.}
        (a) Columns show the RGB input, ground-truth layout, and predictions
        from MIDI~\cite{huang2025midi}, SceneGen~\cite{meng2025scenegen},
        SAM3D~\cite{sam3dteam2025sam3d3dfyimages}, and Bri3D. Red wireframe
        boxes mark the ground-truth object extents, making displacement,
        missing instances, and shape deviations visible. Bri3D better follows
        the reference arrangements in these examples.
        (b) Red rectangles highlight our failures: nearby objects may merge
        in the cluttered seating area (Case 1), and some instances may be
        omitted in the densely populated classroom (Case 2).
    }
    \label{fig:result2}
\end{figure}
\begin{table}[!t]
\caption{
        \textbf{Ablation studies of Bri3D.}
        Rows (b)--(f) progressively introduce the components of the
        Scene Scaffold Generator. Rows (g)/(h) compare interpolation and
        latent refinement from (f); (i)/(j) apply detail synthesis after (h).
        All retain the anchor boxes and BBox IoU of (f). CD-S/F-S and CD-O/F-O report scene/object metrics.
        CD is scaled by $100$, F-Score is in percent, and BBox IoU is in $[0,1]$.
        The best results within each component block are highlighted in bold.
    }
\label{tab:AblationStudyImproved}
    \centering
    \small
    \setlength{\tabcolsep}{3.5pt}
    \renewcommand{\arraystretch}{1.08}

    \resizebox{\linewidth}{!}{%
    \begin{tabular}{c l c c c}
        \toprule
        & \textbf{Variant}
        & \textbf{CD-S/F-S} $\downarrow/\uparrow$
        & \textbf{CD-O/F-O} $\downarrow/\uparrow$
        & \textbf{BBox IoU} $\uparrow$ \\
        \midrule

        \multicolumn{5}{l}{\textit{Baseline}} \\

        (a) & TRELLIS-FT ($64^3$)
            & 10.80/52.91
            & --
            & -- \\

        \midrule
        \multicolumn{5}{l}{\textit{Scene Scaffold Generator}} \\

        (b) & One-hot representation + CE loss ($64^3$)
            & 13.00/43.24
            & 13.90/41.23
            & 0.3144 \\

        (c) & Color representation + $\ell_1$ loss ($64^3$)
            & 7.24/55.57
            & 8.61/53.55
            & 0.3902 \\

        (d) & + Rendering-based label loss
            & 5.47/69.88
            & 7.32/60.82
            & 0.4653 \\

        (e) & + CAG
            & 4.58/73.49
            & 6.35/63.98
            & 0.5804 \\

        (f) & + Higher scaffold resolution ($128^3$)
            & \textbf{3.73/84.22}
            & \textbf{4.33/72.13}
            & \textbf{0.7540} \\

        \midrule
        \multicolumn{5}{l}{\textit{Anchored Voxel Refiner}} \\

        (g) & + Interpolation
            & 1.92/93.53
            & 2.02/91.38
            & \textbf{0.7540} \\
        (h) & + Latent refinement
            & 1.24/94.27
            & 1.85/93.61
            & \textbf{0.7540} \\
        (i) & + Detailed synthesis by Direct3D-S2
            & 0.96/95.82
            & 1.79/95.05
            & \textbf{0.7540} \\
        (j) & + Detailed synthesis by TRELLIS
            & \textbf{0.87/96.12}
            & \textbf{1.72/95.33}
            & \textbf{0.7540} \\

        \bottomrule
    \end{tabular}%
    }

\end{table}

\subsection{Scene Scaffold Generation}
A well-structured spatial layout is crucial for complete and realistic 3D scene generation. Our method supports independent layout prediction.
To evaluate the effectiveness of our Scene Scaffold Generator $\mathcal{G}_S$, we report both quantitative and qualitative results in Table~\ref{tab:ObjectCountIOUTime} and Fig.~\ref{fig:result2} (a), respectively. As shown in Table~\ref{tab:ObjectCountIOUTime}, our $G_S$ achieves a BBox IoU of 0.6732 in the 11--20 object range, significantly outperforming SAM3D~\cite{sam3dteam2025sam3d3dfyimages} and MIDI~\cite{huang2025midi}.
In terms of efficiency, Bri3D produces a scene layout directly
through scene scaffold prediction, taking only 5.1~s on a
NVIDIA A800 GPU for scenes with 1--10 objects. The scaffold already
provides the coarse geometry, instance organization, and spatial
reference needed for object refinement. Subsequent object refinement and detail
synthesis require an additional 37.5~s on average, bringing
the total reconstruction time to 42.6~s per scene.
For full scene reconstruction, Bri3D is approximately
$3.2\times$ faster than SceneGen (135.0~s) and $1.5\times$
faster than SAM3D (64.1~s). For scenes with 11--20 objects,
scaffold prediction takes 5.2~s and full generation takes 78.4~s,
compared with 230.2~s for SceneGen and 179.6~s for SAM3D.
As shown in Fig.~\ref{fig:result2}(b), Bri3D exhibits two
failure modes: (1) under extreme clutter or ambiguous layouts,
it may incorrectly merge distinct objects; and (2) in highly
dense scenes with more than 30 objects, it may omit some
instances.

Qualitative comparisons in Fig.~\ref{fig:result2} (a) further demonstrate that scene scaffold produces visually compact and semantically consistent arrangements, with objects placed in plausible positions relative to room semantics and geometry.
It effectively avoids common artifacts such as excessive gaps between furniture and unrealistic overlaps, which are frequently observed in the outputs of SceneGen~\cite{meng2025scenegen} and MIDI~\cite{huang2025midi}.

\subsection{Ablation Study}
\label{sec:ablation}

We evaluate the contributions of the Scene Scaffold Generator and the
Anchored Voxel Refiner in
Table~\ref{tab:AblationStudyImproved}.

\textbf{Direct object-to-scene transfer.}
We first fine-tune the voxel-generation stage of TRELLIS directly on
scene data (a). Despite its strong object prior, TRELLIS-FT performs
poorly at the scene level, confirming that directly extending an
object-centric generator to full scenes cannot effectively preserve
multi-object geometry and spatial organization.

\textbf{Scene Scaffold Generator.}
Rows (b)--(f) progressively evaluate the scaffold design. Replacing
one-hot labels with color-coded instance representations substantially
improves both scene- and object-level reconstruction, as shown by the
improvement from (b) to (c). Introducing the rendering-based label loss
(d) further preserves instance boundaries and spatial structure,
reducing CD-S from 7.24 to 5.47 and increasing F-S from 55.57
to 69.88. CAG-Attn (e) improves the integration of image and instance
conditions, yielding more coherent object arrangements. Increasing the
scaffold resolution from $64^3$ to $128^3$ (f) provides the strongest
scaffold results, reaching 3.73/84.22 in CD-S/F-S and a BBox IoU of
0.7540.

\begin{figure}[!t]
    \centering
    \includegraphics[width=0.95\linewidth]{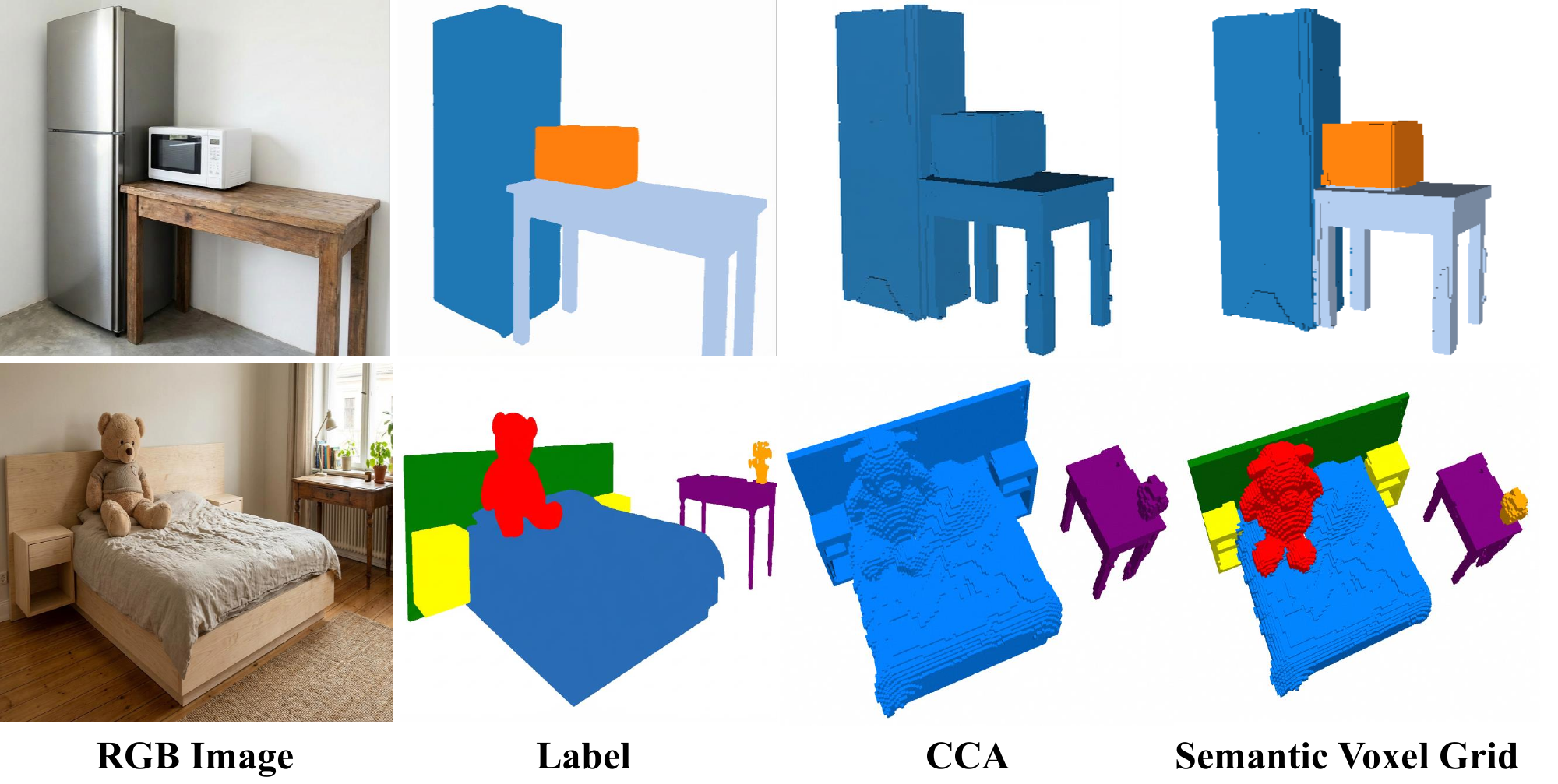}
    \caption{\textbf{Color-coded scene scaffold vs.\ connected component analysis.}
    From left to right: RGB input, instance label map, geometry-only CCA
    result, and the predicted color-coded voxel grid. In the upper example,
    CCA groups the refrigerator, microwave, and table into one connected
    region; in the lower example, it merges the touching bed components
    and teddy bear. The predicted instance colors provide additional
    separation cues at these contacts, allowing joint position--color
    clustering to recover distinct object anchors. Colors indicate instance
    identity rather than surface texture.}
    \label{fig:semantic}
\end{figure}

\textbf{Instance separation with color-coded scene anchors.}
To further demonstrate the effectiveness of the color-coded scene
scaffold, we analyze its critical role in instance separation. As shown
in Fig.~\ref{fig:semantic}, the instance color labels predicted by the
Scene Scaffold Generator $G_S$ enable reliable separation of closely
adjacent objects into distinct instances. In contrast, traditional
geometry-based methods such as connected component analysis (CCA) often
fail to distinguish touching objects in the absence of color guidance.
In the second stage, the Anchored Voxel Refiner $R_A$ leverages the
instance-wise voxel clusters from $G_S$ as strong geometric constraints
to refine object shapes under image conditioning, without relying on
additional instance predictions. Together, this design enables
high-quality 3D scene generation without requiring training from
scratch or large-scale scene datasets.

\textbf{Anchored Voxel Refiner.}
Using configuration (f), we first compare direct voxel interpolation
(g) with scene-anchored latent refinement (h). Although interpolation
increases voxel resolution, it cannot recover missing geometry or align
the coarse scaffold with the latent distribution of the pre-trained
object generator. In contrast, latent refinement improves CD-O/F-O
from 2.02/91.38 to 1.85/93.61, recovering more complete object
structures within the anchor-defined spatial reference. BBox IoU
remains 0.7540 through the retained scaffold-derived anchor boxes.
Finally, rows (i) and (j) pass the high-resolution structure from (h)
to the subsequent generation stages of Direct3D-S2 and TRELLIS,
respectively. Both further improve geometry
without modifying the scaffold-derived anchor boxes. TRELLIS achieves
the best overall performance, with 0.87/96.12 CD-S/F-S,
1.72/95.33 CD-O/F-O, and a BBox IoU of 0.7540, showing that Bri3D can
effectively transfer different pre-trained object priors through the
same scene-anchor interface.

\begin{figure}[!t]
    \centering
    \includegraphics[width=0.95\linewidth]{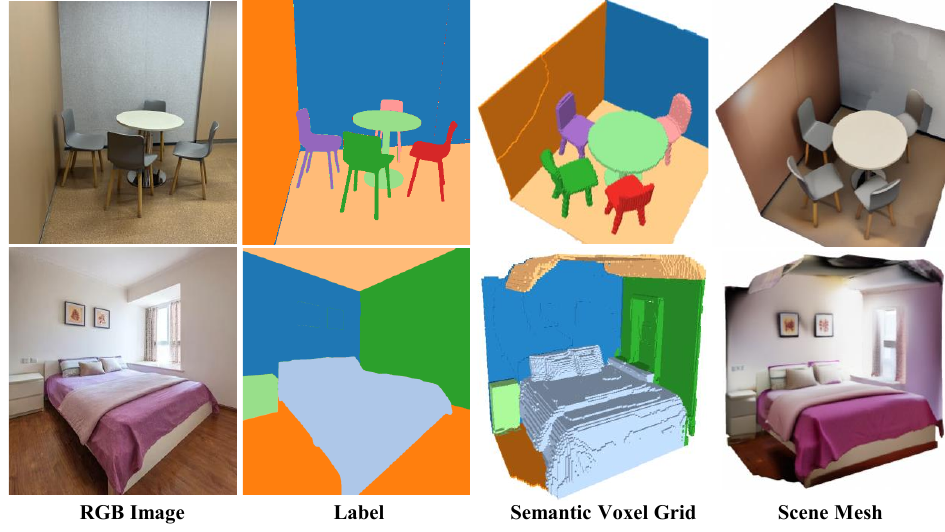}
    \caption{\textbf{Preliminary attempt on structural elements.}
    From left to right: RGB input, label map including structural regions,
    predicted color-coded voxel grid, and reconstructed scene mesh. The
    dining-area and bedroom examples include room surfaces together with
    foreground furniture, illustrating how structural labels can extend
    the scaffold beyond object instances. This extension depends on
    reliable labels for those regions and remains preliminary; floors,
    walls, and ceilings are excluded from the main Bri3D experiments and
    quantitative results.}
    \label{fig:floor_wall}
\end{figure}

\textbf{Preliminary attempt on structural elements.}
Note that Bri3D is designed for foreground objects, and floors, walls,
and ceilings are excluded from reconstruction in all experiments
reported above. Here we only make a \emph{preliminary
attempt} to extend the color-coded scaffold to such large-scale
structural elements. As shown in Fig.~\ref{fig:floor_wall}, when
reliable labels of structural regions are available in the input image,
the scaffold generator can produce coherent floor and wall structures,
indicating the potential of the scene-anchor representation for modeling
global scene structures. A systematic treatment of structural elements,
\eg{} training with clean structural annotations or incorporating
explicit priors such as Manhattan constraints, is left as future work.

\begin{figure}[!t]
    \centering
    \includegraphics[width=0.95\linewidth]{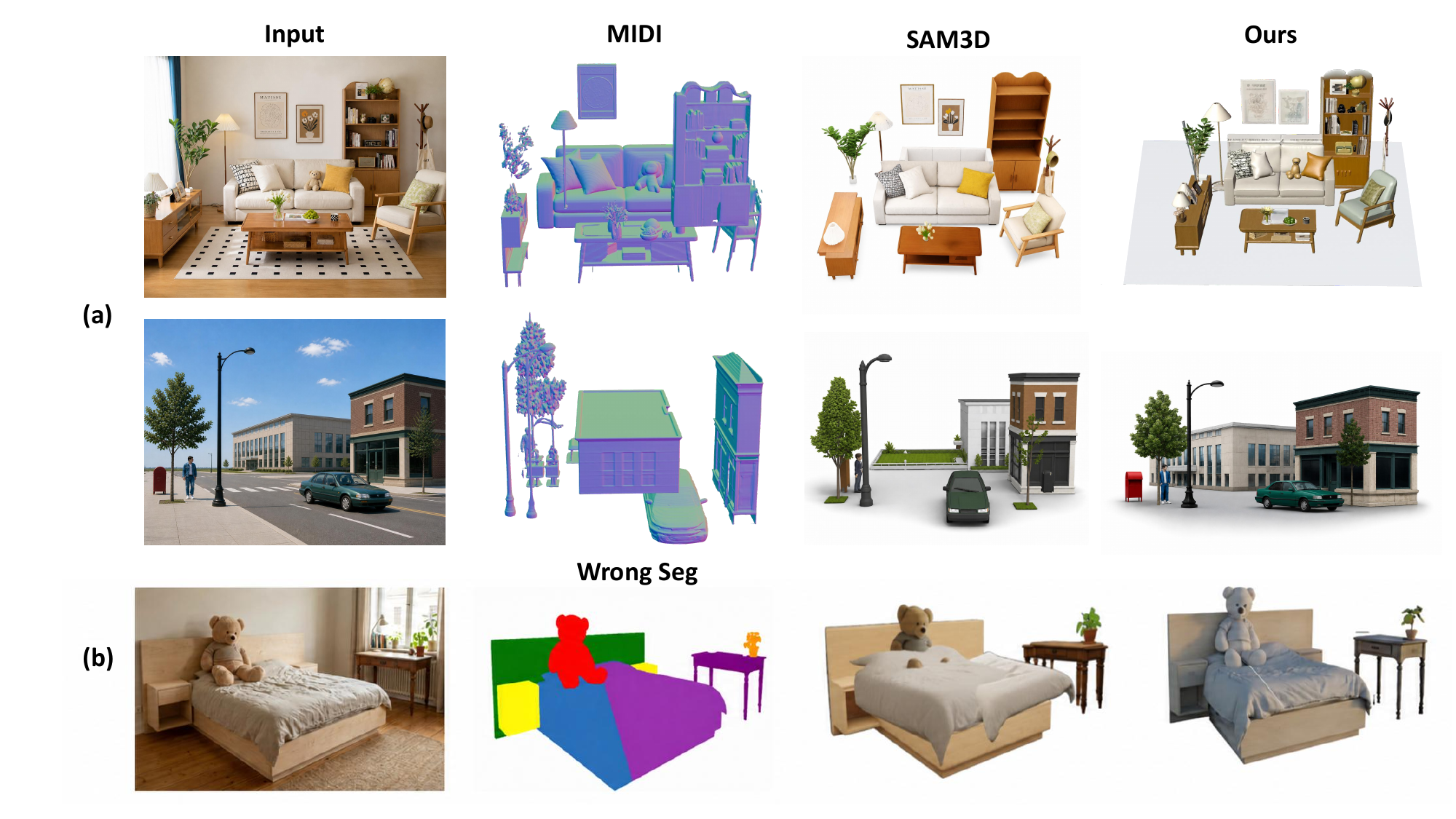}
    \caption{\textbf{Additional qualitative comparisons and robustness to imperfect segmentation.} (a) Geometry and layout comparisons with MIDI and SAM3D on indoor and outdoor scenes; MIDI is shown with normal shading. Bri3D produces coherent layouts and faithful object geometry, while MIDI suffers from distorted shapes and implausible arrangements. (b) With an erroneous instance label map (Wrong Seg), Bri3D exhibits only local cracks, whereas SAM3D produces incorrect object partitions and collisions, including interpenetration between the teddy bear and the bed.}
    \label{fig:more_scenes}
\end{figure}

\textbf{Additional qualitative results and robustness to erroneous segmentation.}
Fig.~\ref{fig:more_scenes}(a) compares Bri3D with MIDI and SAM3D on
both indoor and outdoor scenes. MIDI produces distorted geometry and
implausible object arrangements, whereas Bri3D generates coherent
layouts and object shapes that better match the input image.
The standard quantitative comparisons use dataset-provided instance
masks; here we also examine imperfect segmentation inputs encountered
in practical deployment.
Fig.~\ref{fig:more_scenes}(b) examines robustness to erroneous instance
label maps. Bri3D exhibits only local cracks, whereas SAM3D produces
incorrect object partitions and collisions: the teddy bear is present
but partially intersects the bed.

\begin{figure}[!t]
    \centering
    \includegraphics[width=0.9\linewidth]{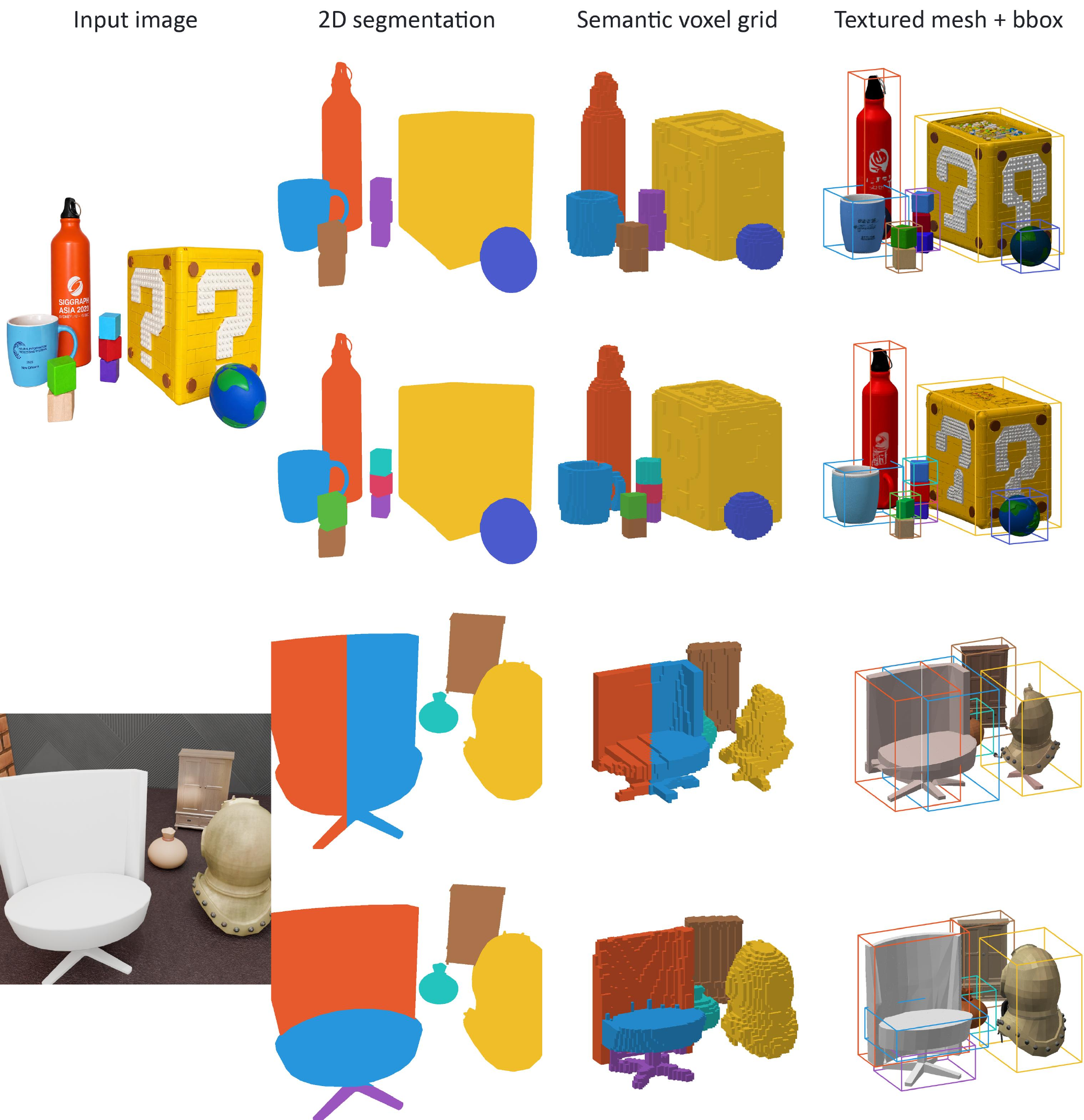}
\caption{
\textbf{Robustness to instance segmentation granularity,
label colors, and implausible partitions.}
From left to right: input image, input instance segmentation,
predicted semantic voxel grid, and reconstructed textured
meshes with bounding boxes.
The first two rows show the same tabletop scene containing a bottle,
a mug, a large box, a ball, and two small stacks of blocks.
The segmentation changes only the grouping of these small stacks:
each stack receives one label in the first row, while its individual
blocks receive separate labels in the second row. 
The last two rows show deliberately implausible partitions
of a chair, together with changes in instance label colors.
Despite these inconsistent instance boundaries, Bri3D
preserves the overall object geometry and spatial arrangement.
}
    \label{fig:more_scenes2}
\end{figure}

\textbf{Robustness to segmentation granularity and label colors.}
Fig.~\ref{fig:more_scenes2} examines how the semantic voxel
grid responds to different instance partitions and label colors.
The first two rows depict a tabletop scene with a bottle, mug,
large box, ball, and two small stacks of blocks. The bottle, mug,
box, and ball keep their instance labels, while the small stacks
are labeled as whole stacks in the first row and as individual
blocks in the second row. The predicted voxel
grids follow these grouping choices while preserving the
overall geometry and spatial arrangement.
The last two rows show implausible partitions of a chair
together with changes in label colors. Although the voxel
partitions and bounding boxes vary with the input labels,
the overall scene geometry and layout remain broadly
consistent. These examples illustrate Bri3D's flexibility
across segmentation granularities and its robustness to
the combined partition and color changes shown.

\section{Conclusion}

We presented Bri3D, a scene-anchored compositional 3D generation framework that achieves coherent spatial layouts and fine-grained object geometry. Bri3D connects scene-level structural reasoning with object-level generative priors through a unified scene scaffold and anchored object refinement. By separating global scene organization from object-scale synthesis, scene-level training focuses on spatial organization and anchor refinement while reusing pretrained priors for detailed geometry and appearance. Through the proposed scene anchors and anchored voxel refinement, Bri3D produces more coordinated and unified scene reconstructions. Experiments demonstrate improvements in scene structure, object geometry, and visual quality. Scene scaffold prediction provides layouts in about five seconds; full scene generation is faster than SceneGen and SAM3D.

\clearpage
\bibliographystyle{assets/plainnat}
\bibliography{main}
\clearpage
\appendix
\setcounter{equation}{0}
\renewcommand{\theequation}{S\arabic{equation}}
\section{Architectural Details}
\label{app:arch}
\subsection{Preliminary: TRELLIS}
\label{subsec:TRELLIS}
TRELLIS~\cite{xiang2025structured} introduces Sparse Structured Latents (SLATs) for efficient 3D generation, representing a shape as $z_{\text{slat}} = \{(z_i, \mathbf{p}_i)\}_{i=1}^L$, where $z_i \in \mathbb{R}^c$ is a latent feature associated with spatial position $\mathbf{p}_i \in \mathbb{R}^3$, and $L \ll N^3$ ensures structural sparsity. The framework adopts a two-stage generation process: sparse voxel generation for coarse object structure, followed by structural latent generation for subsequent decoding into diverse 3D representations such as Meshes, NeRFs~\cite{mildenhall2021nerf} and 3D Gaussians~\cite{kerbl20233d}.

In the first stage, TRELLIS learns a compact representation of sparse voxel structures using a geometry VAE. Given occupied voxels $\{\mathbf{p}_i\}$, we first embed them into a dense occupancy grid $V$, which serves as input to the encoder $E_{\text{geo}}$. The decoder $D_{\text{geo}}$ reconstructs the structure as an occupancy field:
\begin{equation}
    z_{\text{geo}} = E_{\text{geo}}(V), \quad \hat{V} = D_{\text{geo}}(z_{\text{geo}}),
\end{equation}
and the final set of occupied voxels is extracted from $\hat{V}$.
A geometry flow transformer $F_{\text{geo}}$~\cite{peebles2023scalable,lipman2022flow} generates $z_{\text{geo}}$ from noise $\epsilon$, conditioned on an input image $I$ and flow time $t$:
\begin{equation}
    F_{\text{geo}}: (\epsilon, t, I) \mapsto z_{\text{geo}},
\end{equation}
enabling structured 3D shape synthesis from 2D inputs.

Given the decoded positions $\{\mathbf{p}_i\}$, structured noise $\epsilon_{\text{app}} = \{(\epsilon_i, \mathbf{p}_i)\}$ is constructed by associating random noise vectors $\epsilon_i$ with each spatial location. A structured-latent flow transformer $F_{\text{app}}$ then predicts $z_{\text{slat}}$:
\begin{equation}
    F_{\text{app}}: (\epsilon_{\text{app}}, t, I) \mapsto z_{\text{slat}}, \quad O = D_{\text{app}}(z_{\text{slat}}),
\end{equation}
where $D_{\text{app}}$ decodes appearance and $O$ is the final output.
Bri3D leverages this two-stage interface by replacing object-level
voxel generation with scene-level scaffold prediction and anchored
voxel refinement, while keeping the detail-synthesis module frozen.

\subsection{Implementation Details}
\label{sec:impl_details}

\subsubsection{Structure-Color VAE Architecture}
The SC-VAE encodes and decodes the four-channel occupancy and
instance-color representation defined in Section~\ref{sec:scaffold_generator}.
Each 3D-FRONT instance receives a unique color, while Objaverse-XL
objects are partitioned into 1--6 K-means clusters with distinct colors.

The four-channel representation encodes geometric occupancy and
instance identity in a shared latent space. The SC-VAE reconstructs both
channels of information, with voxel reconstruction supervising the 3D
volume and the rendering loss supervising projected instance boundaries.
This representation supplies the scaffold generator with a spatially
organized target that retains the distinction between touching objects.

The core of our generative stages employs a DiT-based architecture. Specifically, the \textit{Scene Scaffold Generator} consists of 24 DiT blocks that jointly process appearance and instance-layout features extracted from the RGB image and its color-coded label map. The Color-Attention Gating module in each cross-attention block adaptively fuses the two conditioning modalities, linking appearance cues with instance-layout evidence.
\subsubsection{Instance Extraction from the Scene Scaffold}
The joint position--color DBSCAN procedure is described in
Section~\ref{sec:refiner}. Its output clusters supply the object anchors
used by the refiner; the training-time K-means labels serve a separate
data-construction purpose.
\subsubsection{Scene-Anchored Latent Refinement}
\label{app:latent_norm}
For the \textit{Anchored Voxel Refiner}, we adopt a rectified flow matching strategy to support enhancement in the latent space. Specifically, the object-scale SC-VAE encoder $E_{\text{SC}}^{o}$ first encodes the coarse object voxel cluster $x_{ci}$ extracted from the scene scaffold into a structural latent, and Gaussian noise is injected to construct the scene-anchored latent initialization $z_i^0 = \mathcal{N}_i[E_{\text{SC}}^{o}(x_{ci})] + \lambda \epsilon_i$ with $\epsilon_i \sim \mathcal{N}(0, I)$. Its encoded component preserves the scaffold-derived pose and coarse geometry, while the noise provides the flexibility required to recover missing and unobserved structures.

For each object, we compute the per-channel mean $\boldsymbol{\mu}_i$
and standard deviation $\boldsymbol{\sigma}_i$ over the spatial positions
of its coarse anchor latent $E_{\mathrm{SC}}^o(\mathbf{x}_i^c)$.
We save these statistics and use the same transformation for the coarse
latent and the corresponding GT target latent:
\begin{equation}
    \mathcal{N}_i(\mathbf{z})=
    \frac{\mathbf{z}-\boldsymbol{\mu}_i}{\boldsymbol{\sigma}_i+\varepsilon},
\end{equation}
where $\varepsilon$ is a small numerical stability constant. Gaussian
noise is added after normalization, as in Eq.~\ref{eq:scene_anchor}. During training, we sample a time step $t \in [0, 1]$ and construct the intermediate latent state $\mathbf{z}(t)$ by interpolating between the scene-anchored initialization $\mathbf{z}_0$ and the target clean latent $\mathbf{z}_1$, \ie{} the GT object latent normalized using the same coarse-anchor statistics:
\begin{equation}
    \mathbf{z}(t) = (1 - t)\,\mathbf{z}_0 + t\,\mathbf{z}_1, \quad t \sim \mathcal{U}[0, 1].
\end{equation}
The model is then trained to predict the velocity field pointing towards the target $\mathbf{z}_1$. The learned flow transforms the scene-anchored initialization into detailed voxel geometry under object image conditioning. The coarse anchor enters through the latent initialization. Before
decoding, we restore the refined latent with the saved statistics:
\begin{equation*}
    \mathcal{N}_i^{-1}(\hat{\mathbf{z}}_i)=
    (\boldsymbol{\sigma}_i+\varepsilon)\odot\hat{\mathbf{z}}_i
    +\boldsymbol{\mu}_i.
\end{equation*}
At inference, these statistics are computed from the input anchor.
\paragraph{Pose preservation and scene composition}
As defined in Section~\ref{sec:refiner}, pose preservation means retaining
the anchor-defined spatial reference. The original center and isotropic
scale are reused in Eq.~\ref{eq:anchor_coordinates} to map each generated
object back into the scene, without additional pose estimation or
registration and without re-normalizing the output bounding box.
Geometry may change within this fixed coordinate reference.
\section{Training Details}
\label{app:training}

\subsection{Data Construction}
We utilize the Objaverse-XL~\cite{deitke2024objaversexl} and 3D-FRONT~\cite{fu20213dfront} datasets to construct our training corpus. Ground truth representations are prepared as \textbf{color-coded voxel grids} to supervise the SC-VAE. Training combines image conditioning with voxel and rendering supervision:
\begin{enumerate}
    \item \textbf{View-Consistent Supervision:} We render ground-truth 3D scenes from the input camera pose to produce the RGB view and its color-coded instance label map. Both images condition the \textit{Scene Scaffold Generator}; only the color-coded instance label map supervises the rendering consistency loss ($\mathcal{L}_{\text{render}}$).
    \item \textbf{Latent Space Supervision:} The 4-channel color-coded voxel grids serve as the direct reconstruction target for the SC-VAE, providing dense supervision for learning the structured latent prior.
\end{enumerate}

The multi-resolution coarse conditions and their $64^3$ encoder inputs
are defined in Section~\ref{sec:refiner}. The object-level corpus size
and scene-level data source are reported separately in
Section~\ref{sec:experimental_setup}.

\subsection{Computational Details}
We implement the models in PyTorch with Distributed Data Parallel
(DDP). Training uses the hardware and hyperparameters reported in
Section~\ref{sec:experimental_setup}. The total training duration is one
week, corresponding to 2,688 GPU-hours, and the model has 555.7 million
parameters. Inference timing in Table~\ref{tab:ObjectCountIOUTime}
is measured on a single NVIDIA A800 GPU. Bri3D uses 25 sampling steps
and processes object instances in parallel.
\section{Additional Evaluation Details}
\label{app:evaluation}

\subsection{Cross-Dataset Evaluation Set}
The scene sources, manual selection, and multi-view input rendering
are described in Section~\ref{sec:experimental_setup}.
Table~\ref{tab:new_dataset_generalization} evaluates the combined set;
multiple views of one scene are distinct single-image reconstruction
inputs.

\subsection{Geometric Evaluation Protocol}
We follow the geometric evaluation settings of
MIDI~\cite{huang2025midi} for Chamfer Distance (CD),
F-Score, and bounding-box IoU (BBox IoU). We use MIDI's evaluation
implementation and settings for point sampling, distance computation,
and instance matching across the compared methods. The F-Score distance
threshold is $0.1$ in the evaluation coordinate system. CD and F-Score are evaluated
at both the scene and object levels, while bounding-box
IoU measures agreement between corresponding axis-aligned
object bounding boxes. For Bri3D, these are the scaffold-derived anchor
boxes retained throughout refinement and detail synthesis, measuring
the layout established by the scaffold. CD and F-Score evaluate the
geometry produced by each stage.
Throughout Tables~\ref{tab:scene_metrics},
\ref{tab:new_dataset_generalization}, and
\ref{tab:AblationStudyImproved}, CD-S and CD-O denote scene- and
object-level CD, respectively, and are reported as $100\times$ the
unscaled distances. F-S and F-O denote scene- and object-level
F-Score and are reported as percentages. BBox IoU is reported on
the $[0,1]$ scale. Collision rates in
Table~\ref{tab:ObjectCountIOUTime} are percentages of overlapping
object pairs, obtained by multiplying the overlap fraction by $100$;
for example, a fraction of 0.04 is reported as $4\%$. For Bri3D,
we extract the scaffold instances and voxelize them separately in the
shared scene coordinate system before refinement. Collision detection compares the resulting instance occupancies
in this common coordinate system.

\end{document}